\documentclass[11pt]{article}
\usepackage{coling2020}
\usepackage{times}
\usepackage{url}
\usepackage{latexsym}
\usepackage{tikz}
\usetikzlibrary{shapes,arrows,chains}
\usepackage{tikz-dependency}
\usepackage{multirow}
\usepackage{array}
\usepackage{subfigure}
\usepackage{float}
\usepackage{setspace}
\usepackage{verbatim}
\usepackage{lingmacros}
\usepackage{forest}
\useforestlibrary{linguistics}
\forestapplylibrarydefaults{linguistics}
\usepackage{linguex}
\usepackage{booktabs}
\usepackage{todonotes}
\usepackage{amsmath}
\usepackage{tikz-qtree}
\usepackage{float}
\usepackage{tikz-qtree-compat}
\usepackage{pgfplots}
\usepackage{caption}
\definecolor{airforceblue}{rgb}{0.36, 0.54, 0.66}
\definecolor{antiquebrass}{rgb}{0.8, 0.58, 0.46}
\definecolor{antiquefuchsia}{rgb}{0.57, 0.36, 0.51}
\definecolor{ao(english)}{rgb}{0.0, 0.5, 0.0}

\colingfinalcopy % Uncomment this line for the final submission

\title{Constructing a Family Tree of Ten Indo-European Languages with Delexicalized Cross-linguistic Transfer Patterns}
  \author{Yuanyuan Zhao$^{1,2}$, Weiwei Sun$^{1,3}$ and Xiaojun Wan$^{1}$\\
    $^1$Wangxuan Institute of Computer Technology, Peking University \\
    $^1$The MOE Key Laboratory of Computational Linguistics, Peking University\\
    $^2$Academy for Advanced Interdisciplinary Studies, Peking University \\
    $^3$Center for Chinese Linguistics, Peking University\\
    \texttt{\{zhao\_yy,ws,wanxiaojun\}@pku.edu.cn} \\}
\date{}

\begin{document}
\maketitle
\begin{abstract}
It is reasonable to hypothesize that the divergence patterns formulated by historical linguists and typologists reflect constraints on human languages, and are thus consistent with Second Language Acquisition (SLA) in a certain way. In this paper, we validate this hypothesis on ten Indo-European languages. We formalize the delexicalized transfer as interpretable tree-to-string and tree-to-tree patterns which can be automatically induced from web data by applying neural syntactic parsing and grammar induction technologies.
This allows us to quantitatively probe cross-linguistic transfer and extend inquiries of SLA.
We extend existing works which utilize mixed features and support the agreement between delexicalized cross-linguistic transfer and the phylogenetic structure resulting from the historical-comparative paradigm.
\end{abstract}

\section{Introduction}

The traditional historical classification of Indo-European languages generally focuses on diachronic language change, based on the cross-lingual patterns of lexical entities and of sound laws connecting cognate words \cite{clackson2007indo}.
%Recently, a series of works propose to address this issue by utilizing various combinations of linguistic features extracted from non-native (L2) texts\footnote{Henceforth, non-native data and its corrected version are referred to as L2 and L1, respectively.} \cite{nagata2013reconstructing,nagata2014language,berzak2014reconstructing,rabinovich2018native}.
Recently, several preliminary works suggest the coherence between such classification and derived results with various combinations of linguistic features extracted from non-native (L2) texts\footnote{Henceforth, non-native data and its corrected version are referred to as L2 and L1, respectively.} \cite{nagata2013reconstructing,nagata2014language,berzak2014reconstructing,rabinovich2018native}.
They demonstrate the hypothesis that cross-linguistic transfer\footnote{This is different from the widely-known \textit{cross-lingual transfer learning} in the NLP community which broadly means adapting models from resource-rich languages to resource-poor languages.}, which refers to the application of learners' knowledge of one language to the use of another as shown in Figure \ref{fig:clt}, is so strong that we can reveal the language family relationships preserved in L2 texts.
The limitation is that they all use mixed features across lexical, syntactic and semantic dimensions.
In view of the fact that \textit{cognates} across languages provide credible genealogical clues, it is not clear to what extent does a single factor reflect the influence from traditional phylogenetic language groups.

Previous works \cite{Dun:Ter:Ree:05,murawaki2015continuous} have investigated the role of syntactic structures in phylogeny prediction, especially from the perspective of modern generative theory \cite{longobardi2009evidence}.
In this paper, we concentrate on the cross-lingually transferred delexicalized feature and aim to find out whether it can work as effectively as lexical information such as \textit{cognates}.

\begin{minipage}[t]{\textwidth}
\centering
\begin{tikzpicture}

    \node(a)[rounded corners,draw=blue] {play};
    \node(b)[rounded corners, draw=blue, right = 0.3 of a] {often};
    \node(c)[rounded corners, draw=blue, right = 0.3 of b] {sports};
    \node(m) [left = 0.2 of a]{\textbf{L2 English}};

    \node (d) [above = 0.3 of a]{faire};
    \node(e) [right = 0.1 of d]{souvent};
    \node(f) [right = 0.1 of e]{du};
    \node(g) [above = 0.3 of c, xshift=0.7cm]{sport};
    \node(h) [left = 0.2 of d]{\textbf{L1 French}};

    \node(i)[rounded corners,  below = 0.3 of a,draw=blue] {often};
    \node(j)[rounded corners,  below = 0.3 of b,draw=blue] {play};
    \node(k)[rounded corners,  below = 0.3 of c,draw=blue] {sports};
    \node(l) [left = 0.2 of i]{\textbf{L1 English}};

    \draw[dashed,thick] (a) to (d);
    \draw[dashed,thick] (b) to (e);
    \draw[dashed,thick] (c) to (g);
    \draw[dashed,thick] (a) to (j);
    \draw[dashed,thick] (b) to (i);
    \draw[dashed,thick] (c) to (k);
\end{tikzpicture}
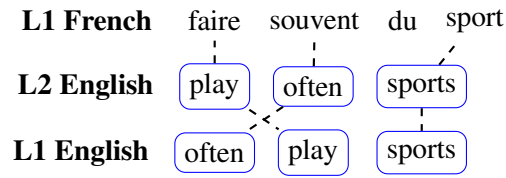
\makeatletter\def\@captype{figure}\makeatother\caption{Example of cross-linguistic transfer. The L2 English phrase is written by a French native speaker and it preserves linguistic features from L1 French, which results in the crossing of \textit{VB} and \textit{ADVP} in Figure \ref{fig:pattern_fr}.}
\makeatletter\def\@captype{figure}\makeatother\label{fig:clt}
\end{minipage}

Based on parallel L2-L1 data, we design two interpretable structures: tree-to-string and tree-to-tree patterns, as illustrated in Figure \ref{fig:pattern}.
Such patterns can be automatically induced from large-scale data by means of state-of-the-art neural syntactic parsing and grammar induction technologies.
In particular, tree-to-string patterns are derived in the framework of Synchronous Tree-Substitution Grammar (STSG; \newcite{zhang2006synchronous}), while tree-to-tree patterns are generated by a novel tree alignment algorithm.
Then we build a large language-pattern matrix, with which we can probe cross-linguistic transfer in a quantitative manner and assist the study of second language acquisition (SLA).
Experiments on ten representative Indo-European languages, %belonging to Romance, Germanic, Slavic, Indic and Iranian families,
show that with the delexicalized patterns we can reproduce the exact phylogenetic structure resulting from the historical-comparative paradigm.

\begin{figure}[htbp]
\centering
\subfigure[Common]{
\begin{minipage}[t]{0.2\linewidth}
\centering
\begin{tikzpicture}
      \tikzset{level distance=20pt}
      \tikzset{every tree node/.style={align=center,anchor=north}}
    \Tree [.NP [.NP \node(a){$x_0$}; ] [.PP \node(b){$x_1$}; ] ];
    \draw[dashed, thick, red] (-1,-2.7) rectangle (1,0) ;
    \draw[dashed, thick, blue] (-1.3,-4.1) rectangle (1.3,0.2) ;
    \begin{scope}[yshift=-3.5cm,grow'=up]
    \Tree [.NP [.NP \node(c){$x_1$}; ] [.NN \node(d){$x_0$}; ] ];
    \end{scope}
    \draw[dotted,thick] (a) to (d);
    \draw[dotted,thick] (b) to (c);
\end{tikzpicture}
\end{minipage}
}
\subfigure[Mandarin]{
\begin{minipage}[t]{0.2\linewidth}
\centering
\begin{tikzpicture}
      \tikzset{level distance=20pt}
      \tikzset{every tree node/.style={align=center,anchor=north}}
    \Tree [.NP [.NN \node(a){$x_0$}; ] [.NN \node(b){$x_1$}; ] ];
    \draw[dashed, thick, red] (-1,-2.7) rectangle (1,0) ;
    \draw[dashed, thick, blue] (-1.3,-4.1) rectangle (1.3,0.2) ;
    \begin{scope}[yshift=-3.5cm,grow'=up]
    \Tree [.NP [.NN \node(c){$x_1$}; ] [.NN \node(d){$x_0$}; ] ];
    \end{scope}
    \draw[dotted,thick] (a) to (d);
    \draw[dotted,thick] (b) to (c);
\end{tikzpicture}
\end{minipage}
}
\subfigure[French]{
\begin{minipage}[t]{0.25\linewidth}
\centering
\begin{tikzpicture}
      \tikzset{level distance=20pt, sibling distance=.02cm}
      \tikzset{every tree node/.style={align=center,anchor=north}}
    \Tree [.VP [.VB \node(a){$x_0$}; ] [.ADVP \node(b){$x_1$}; ] [.NP \node(c){$x_2$}; ] ];
    \draw[dashed, thick, red] (-1.5,-2.7) rectangle (1.5,0) ;
    \draw[dashed, thick, blue] (-1.8,-4.8) rectangle (1.8,0.2) ;
    \begin{scope}[xshift=-0.2cm,yshift=-4.2cm,grow'=up]
    \Tree [.VP [.ADVP \node(d)[above = 0.2 of a]{$x_1$}; ] [.VP [.VB \node(e){$x_0$}; ] [.NP \node(f){$x_2$}; ] ] ];
    \end{scope}
    \draw[dotted,thick] (a) to (e);
    \draw[dotted,thick] (b) to (d);
    \draw[dotted,thick] (c) to (f);
\end{tikzpicture}
\makeatletter\def\@captype{figure}\makeatother\label{fig:pattern_fr}
\end{minipage}
}
\subfigure[Arabic]{
\begin{minipage}[t]{0.2\linewidth}
\centering
\begin{tikzpicture}
      \tikzset{level distance=20pt, sibling distance=.02cm}
      \tikzset{every tree node/.style={align=center,anchor=north}}
    \Tree [.S [.NP \node(a){$x_0$}; ] [.VP [.MD \node(b){$x_1$}; ] [.VP \node(c){$x_2$}; ] ] ];
    \draw[dashed, thick, red] (-1.1,-3.4) rectangle (1.5,0) ;
    \draw[dashed, thick, blue] (-1.4,-4.8) rectangle (1.8,0.2) ;
    \begin{scope}[xshift=0.2cm,yshift=-4.2cm,grow'=up]
    \Tree [.SQ [.MD \node(d){$x_1$}; ] [.NP \node(e){$x_0$}; ] [.VP \node(f){$x_2$}; ] ] ];
    \end{scope}
    \draw[dotted,thick] (a) to (e);
    \draw[dotted,thick] (b) to (d);
    \draw[dotted,thick] (c) to (f);
\end{tikzpicture}
\end{minipage}
}
\caption{Examples of cross-lingual transfer patterns summarized as tree-to-string (red box) and tree-to-tree (blue box) patterns.
The top and bottom trees correspond to the syntactic analysis for L2-English and their corrected sentences respectively.
$x_0$, $x_1$ and $x_2$ represent constituent tokens, among which the dashed lines refer to word alignment.
(a) depicts a top-ranking pattern for all native languages and
(b)--(d) show remarkably frequent patterns related to language-specific properties of different native languages:
(b) reflects the heavy use of noun-noun compound in Mandarin Chinese;
(c) reflects the different word orders in English and French related to adverbial adjuncts;
(d) reflects the different word orders in English and Arabic related to Modal words.
}
\label{fig:pattern}
\end{figure}
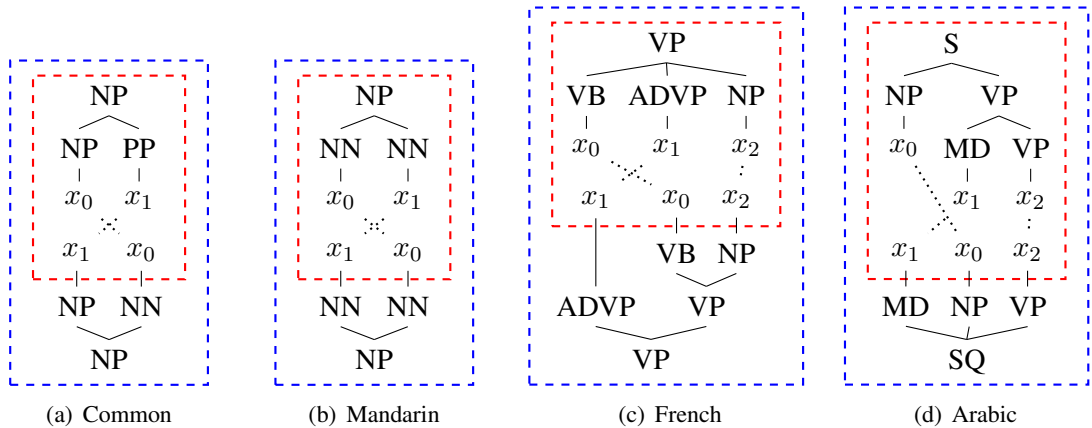

\section{Inducing Delexicalized Cross-lingual Transfer Patterns}
\label{sec:pattern}
We define two kinds of syntactically motivated structures (tree-to-string and tree-to-tree patterns) to measure the inherent transfer, as illustrated in Figure \ref{fig:pattern}.
Both of them intuitively show structural language variances, largely word order errors.
Based on reliable syntactic analysis for aligned parallel data\footnote{The alignment between L2 and L1 tokens is calculated with {\tt BerkeleyAligner} (\url{https://code.google.com/archive/p/berkeleyaligner/}).}, we can generate such patterns with grammar induction technologies.

\subsection{Inducing Tree-to-string Patterns with STSG}
Tree-substitution grammar (TSG) defines the model of deriving parse trees with a rewriting system and has been proved effective to capture native language signals \cite{swanson2013extracting}.
Here we propose to utilize Synchronous TSG (STSG) which describes the process of gathering tree-to-string fragments from aligned hierarchical (L2 trees) and flat structures (L1 strings).
%because it can explicitly represent the non-canonical expressions which largely result from cross-linguistic transfer.
We derive such patterns with a modified GHKM algorithm \cite{galley2004s}.
%, an instantiation of STSG.
In particular, reordering across all levels of a parse tree makes it possible to capture the divergence of word orders.
%which is a vital manifestation of syntactic transfer.
Note that we only keep reordering rules and exclude patterns containing any lexical information.
\subsection{Inducing Tree-to-tree Patterns with Tree Alignment Algorithm}
We design a novel alignment algorithm to induce tree-to-tree patterns which can provide richer structural information.
Let $T_1$ and $T_2$ denote the constituent trees for L1 and L2 sentences respectively.
For each subtree $T^k_2$ in $T_2$, suppose that it corresponds to the span [$w^b_2$, $w^e_2$] in the L2 sentence and the indexes of their mapped tokens in the L1 sentence are $p^b_1$, ..., $p^e_1$.
$T^k_2$ is regarded as \textit{anchor} iff there exist $i$ and $j$, satisfying the following conditions:
$$b\leq i < j \leq e, p^i_1 > p^j_1.$$
These \textit{anchors} indicate token-level reordering where our algorithm starts.
Their aligned trees in $T_1$ are defined as the minimum subtree which covers $p^b_1$, ..., $p^e_1$.
Then we can extract a tree-to-tree pattern by removing lexical information.
To avoid repeated extraction, we carry out the algorithm in a bottom-up recursion and record the visited subtrees during the process.
With this algorithm, we can induce the minimalist pattern set for each sentence pair efficiently.

%
%%\begin{minipage}{\textwidth}

%%\begin{minipage}[b]{0.45\textwidth}
%%  \begin{tikzpicture}
%%      \tikzset{level distance=24pt}
%%      \tikzset{every tree node/.style={align=center,anchor=north}}
%%    %\Tree [.Indo-European [ [.Indic Hindi ] [.Iranian Persian ] ]   relies upon vocabulary items only
%%    %                        [.Slavic Polish [ Ukrainian Russian ] ]
%%    %                        [ [.Germanic German ] [.Romance Italian [ French [ Spanish Portuguese ] ] ] ] ]
%%    \Tree [.Indo-European [.Indic Hin ] [.Iranian Per ] [.Slavic Pol Ukr Rus ] [.Romance Ita Fre Spa Por ] ]
%%    \end{tikzpicture}
%%  \makeatletter\def\@captype{figure}\makeatother\caption{Indo-European family tree.}
%%  \makeatletter\def\@captype{figure}\makeatother\label{fig:ietree}
%%\end{minipage}
%%\end{minipage}
%
%

\section{Experiment}
\subsection{Data}
\label{sec:data}
%Recent years have seen an increasing interest in the exploration of learner language by investigating non-native texts in a systematic manner.
%Several projects have been proposed which aim at establishing learner corpora by firstly formulating an annotation guideline \cite{ragheb:dickinson:12,nagata2016phrase,berzak2016universal}.
%There are also two large-scale corpora of L2 English sentences, collected from social networking sites (SNSs), one for high-fluent English collected from Reddit\footnote{\url{https://www.reddit.com/}} \cite{rabinovich2018native} and the other for intermediate English from Lang-8\footnote{\url{https://lang-8.com/}} \cite{mizumoto2011mining}.

Following \newcite{mizumoto2011mining}, we extract a large volume of L2-L1 sentence pairs by mining the revision logs of authors who are learning English on Lang-8\footnote{\url{https://lang-8.com/}}.
The initial sentences written by learners can be regarded as L2 texts, while the sentences corrected by native speakers can be regarded as L1 texts.
After careful data cleansing, we build a corpus of 6,791,165 parallel sentence pairs from writers of 96 different native languages, and the corpus covers a wide variety of topics and genres.
We keep 21 languages each of which has at least 10,000 sentence pairs in the corpus for pattern analysis and 10 of them are in the Indo-European language family which we use as our experimental data.

\newcommand{\tabincell}[2]{\begin{tabular}{@{}#1@{}}#2\end{tabular}}
\begin{minipage}{\textwidth}
%\begin{minipage}[c]{0.5\textwidth}
%\centering
%\begin{tabular}{c|c|c}
%		\toprule
%		\tabincell{c}{Treebanks}& \tabincell{c}{\#Word/Sent} &\tabincell{c}{\#Sent}\\
%		\midrule
%		\tabincell{l}{WSJ 02-21} & \tabincell{c}{23.85} &\tabincell{c}{39832} \\
%		\tabincell{l}{ICNALE} & \tabincell{c}{17.62} &\tabincell{c}{1930} \\
%		\tabincell{l}{Konan-JIEM} & \tabincell{c}{9.42} &\tabincell{c}{3260}\\
%		\bottomrule
%	\end{tabular}
%\vspace{5mm}
%\makeatletter\def\@captype{table}\makeatother\caption{Statistics of treebanks.}
%\makeatletter\def\@captype{table}\makeatother\label{tab:statistics}
%\end{minipage}

\begin{minipage}[c]{0.4\textwidth}
\centering
	\begin{tikzpicture}
\centering
\begin{axis}[height=5cm,width=8.5cm,xtick pos=left,xtick=data,
xticklabels={0,1K,2K,3K,3.39K},
ytick pos=left,legend pos=south east,legend style={draw=none}]
\addplot[smooth,color=red,mark=triangle] coordinates {
(0,89.43)
(1,92.31)
(2,92.81)
(3,93.11)
(3.39,92.98)
};
\addlegendentry{\textit{$F_1$}}
\end{axis}
\end{tikzpicture}
\vspace{-5mm}
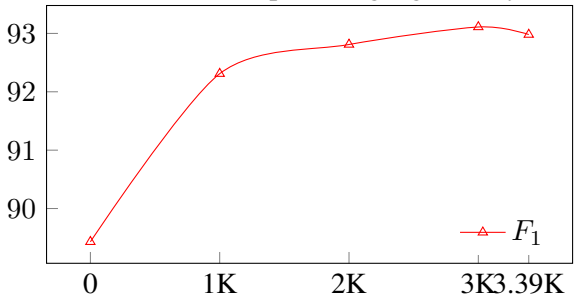
\makeatletter\def\@captype{figure}\makeatother\caption{Parsing performance with incrementally added L2 trees.}
\makeatletter\def\@captype{figure}\makeatother\label{fig:parsing}
\end{minipage}
\hspace{2cm}
\begin{minipage}[c]{0.4\textwidth}
\centering
\begin{tabular}{c|c|c}
  \hline
  Feature & \tabincell{c}{Purity\\ Score}&\tabincell{c}{Leaf-pair\\ Distance}\\
  \hline
  Word Pairs & 100 & 8.11\\
   CFG Rules & 70.74 &11.40\\
   Tree-to-string Patterns& 100&6.42\\
  Tree-to-tree Patterns&100&7.82 \\
  \hline
\end{tabular}
\vspace{5mm}
\makeatletter\def\@captype{table}\makeatother\caption{Evaluation results of clustered trees based on different features.}
\makeatletter\def\@captype{table}\makeatother\label{tab:res}
\end{minipage}
\end{minipage}
\subsection{Tailoring Parser to L2}
L2 texts exhibit many distinguishing features compared to their L1 counterparts, but almost all existing parsers are fine-tuned on L1s.
We propose to tailor parsing models to L2 by incrementally enhancing the standard parser with learner data.
%, which can not only drag the feature distribution closer to the test space but also improve the generalization ability.
We choose the Berkeley Neural Parser \cite{kitaev2018multilingual} which incorporates self-attention layers \cite{kitaev2018constituency} and BERT word representations \cite{devlin2018bert}.
The original training data consists of trees from WSJ 02-21 sections of Penn TreeBank (PTB; \newcite{ptb}).
L2 trees are extracted from two learner English treebanks, %manually annotated with phrase structures,
i.e., Konan-JIEM and ICNALE \cite{nagata2016phrase}.
%Statistics are shown in Table \ref{tab:statistics}.
We extract 3390, 1000 and 800 L2 trees from these two treebanks in proportion as the supplementary-training, development and test sets respectively.

Parsing results are displayed in Figure \ref{fig:parsing}.
%While adding supplementary L2 trees into the training data incrementally, a great improvement can be observed at beginning, but the increase gets smaller with more data added.
The overall trend indicates that expanding training data can bring about a positive, but limited effect.
The best-performing model can achieve an $F_1$ score of 93.11\%, which shows that the tailored parser can produce relatively reliable syntactic trees for L2 English sentences.

\subsection{Hierarchical Clustering}

To facilitate further retrieval and calculation, we build a large language-pattern matrix with dimensions of $N_p \times N_l$, where $N_p$ and $N_l$ refer to the total numbers of extracted patterns and native languages respectively, and utilize principal component analysis (PCA) to reduce dimensionality.
% The values are calculated by dividing the observed frequency by the total number of sentences of the current language.
% Due to the sparsity, we utilize principal component analysis (PCA) to reduce dimensionality.
% The resulted matrix ($N_r \times N_l$) illustrates the pattern distribution and can serve as a measure of language mastery and challenges by speaker’s native language.
% In addition, we can induce the relative values of interpretable typological features from the matrix, like a subset of word order features from WALS \cite{wals}.
% Each language is encoded with a dense embedding so we can directly calculate the similarity between any two languages.
Based on the matrix, we obtain a square matrix with dimensions of $N_l \times N_l$ by calculating the Euclidean distance between any two languages.
Then we use hierarchical clustering \cite{mullner2011modern,mullner2013fastcluster} to produce the dendrogram.
%The algorithm starts with a pile of clusters, each containing a single item corresponding to a language.
%Then we repeatedly combine the closest two items until the formation of a whole cluster.
%To test the robustness of the proposed pattern,
%we try several different distance functions and clustering methods. (See Appendices for more details.)

%\section{Experiment and Discussion}
\subsection{Evaluation}
Several methodologies have been proposed for evaluating phylogenetic trees \cite{robinson1981comparison,kuhner1994simulation,heller2005bayesian,teh2008bayesian,rabinovich2017found}.
We choose purity score \cite{heller2005bayesian} and leaf-pair distance \cite{rabinovich2017found} as our evaluation metric.
The first method finds the smallest subtree containing leaves with the same discrete class and the purity is 100\% iff all leaves in each class are contained in some pure subtree.
The second method calculates the average length of shortest paths in terms of any two leaves.
A smaller distance represents higher structural similarity.
%Several methodologies have been proposed for evaluating phylogenetic trees, including Robinson-Foulds \cite{robinson1981comparison}, branch score \cite{kuhner1994simulation}, purity score \cite{heller2005bayesian}, subtree score \cite{teh2008bayesian} and leaf-pair distance \cite{rabinovich2017found}.
% Several methodologies have been proposed for evaluating phylogenetic trees \cite{robinson1981comparison,kuhner1994simulation,heller2005bayesian,teh2008bayesian,rabinovich2017found}.
% We choose purity score \cite{heller2005bayesian} as our evaluation metric which focuses on structural similarities.
% Let $T_g$ be a tree with $n$ leaves  whose discrete class labels are $c_1$, . . . , $c_n$.
% For each leave pair $i$ and $j$ from the same class, i.e. $c_i$ = $c_j$, find the smallest subtree containing $i$ and $j$ in another tree.
% Measure the fraction of leaves in that subtree which are in the same class ($c_i$).
% The expected value of this fraction is the dendrogram purity.
% The purity is 1 iff all leaves in each class are contained in some pure subtree.
\subsection{Result and Analysis}

%The second pattern can be attributed to the crossing of verbs and adverbs between French and English: adverbs in French generally follows the verb which is less common in English.
%The second one corresponds to the fact that Arabic word orders normally do not change even in interrogative sentences which instead in English is typically marked by inversion of the subject and predicate.
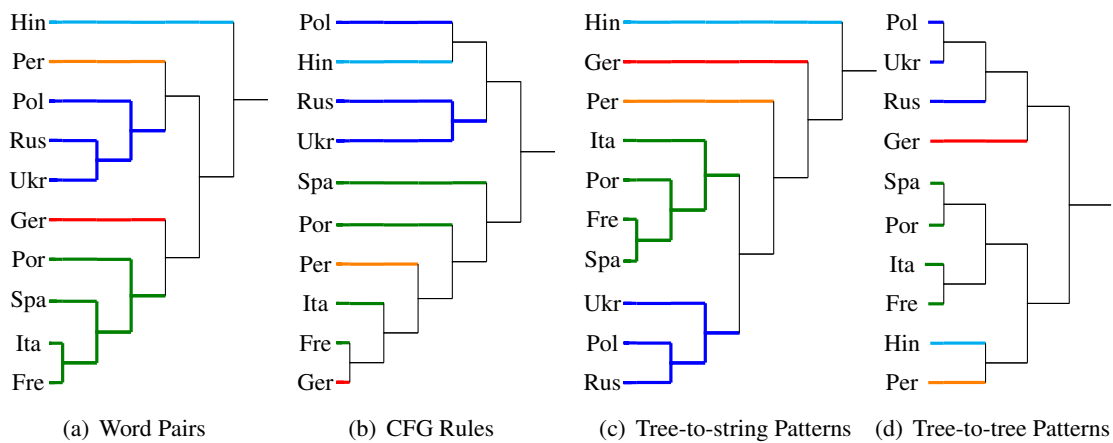
\begin{figure*}
\centering
\begin{minipage}[b]{0.23\textwidth}
	\subfigure[Word Pairs]{
	\begin{tikzpicture}[every tree node/.style={font=\small},
            grow=left,
            level distance=0.45cm,sibling distance=.05cm,
            edge from parent path={(\tikzparentnode.south) -- +(-5pt,0) -| (\tikzchildnode)}],
            \Tree[
            \edge node[near end,left] {};
            [
            \edge[cyan,very thick] node[] {}; [\edge[cyan,very thick]; [\edge[cyan,very thick]; [\edge[cyan,very thick]; [\edge[cyan,very thick]; [\edge[cyan,very thick]; [. Hin ]]]]]]
            ]
            \edge node[near end,right] {};
            [
            \edge node[near end,left] {};
            [
            \edge node[near end,left] {};
            [
            \edge[orange,very thick] node[] {}; [\edge[orange,very thick]; [\edge[orange,very thick]; [\edge[orange,very thick]; [. Per ]]]]
            ]
            \edge node[near end,right] {};
            [
            \edge[blue,very thick] node[near end,left] {};
            [
            \edge[blue,very thick] node[] {}; [\edge[blue,very thick]; [\edge[blue,very thick]; [. Pol ]]]
            ]
            \edge[blue,very thick] node[near end,right] {};
            [
            \edge[blue,very thick] node[near end,left] {};
            [
            \edge[blue,very thick] node[] {}; [\edge[blue,very thick]; [. Rus ]]
            ]
            \edge[blue,very thick] node[near end,left] {};
            [
            \edge[blue,very thick] node[] {}; [\edge[blue,very thick]; [. Ukr ]]
            ]]]]
            \edge node[near end,right] {};
            [
            \edge node[near end,left] {};
            [
            \edge[red,very thick] node[] {}; [\edge[red,very thick]; [\edge[red,very thick]; [\edge[red,very thick]; [. Ger ]]]]
            ]
            \edge node[near end,right] {};
            [
            \edge[ao(english),very thick] node[near end,left] {};
            [
            \edge[ao(english),very thick] node[] {}; [\edge[ao(english),very thick]; [\edge[ao(english),very thick]; [. Por ]]]
            ]
            \edge[ao(english),very thick] node[near end,right] {};
            [
            \edge[ao(english),very thick] node[near end,left] {};
            [
            \edge[ao(english),very thick] node[] {}; [\edge[ao(english),very thick]; [. Spa ]]
            ]
            \edge[ao(english),very thick] node[near end,right] {};
            [
            \edge[ao(english),very thick] node[near end,left] {};
            [
            \edge[ao(english),very thick] node[] {}; [. Ita ]
            ]
            \edge[ao(english),very thick] node[near end,right] {};
            [
            \edge[ao(english),very thick] node[] {}; [. Fre ]
            ]] ]]] ]]
            \end{tikzpicture}
            }
\end{minipage}
\begin{minipage}[b]{0.23\textwidth}
	\subfigure[CFG Rules]{	
            \begin{tikzpicture}[every tree node/.style={font=\small},
            grow=left,
            level distance=0.45cm,sibling distance=.05cm,
            edge from parent path={(\tikzparentnode.south) -- +(-5pt,0) -| (\tikzchildnode)}],
            % frontier/.style={distance from root=50pt} % Align leaf nodes
            \Tree
            [
            \edge node[near end,left] {};
            [
            \edge node[near end, left] {};
            [\edge node[near end,left] {};
            [
            \edge[blue,very thick] node[] {}; [\edge[blue,very thick]; [\edge[blue,very thick]; [\edge[blue,very thick]; [. Pol ]]]]
            ]
            \edge node[near end,right] {};
            [
            \edge[cyan,very thick] node[] {}; [\edge[cyan,very thick]; [\edge[cyan,very thick]; [\edge[cyan,very thick]; [.Hin ]]]]
            ]
            ]
            \edge node[near end,right] {};
            [\edge[blue,very thick] node[near end,left] {};
            [
            \edge[blue,very thick] node[] {}; [\edge[blue,very thick]; [\edge[blue,very thick]; [\edge[blue,very thick]; [.Rus ]]]]
            ]
            \edge[blue,very thick] node[near end,right] {};
            [
            \edge[blue,very thick] node[] {}; [\edge[blue,very thick]; [\edge[blue,very thick]; [\edge[blue,very thick]; [.Ukr ]]]]
            ]
            ]
            ]
            \edge node[near end,right] {};
            [
            \edge node[near end,left] {};
            [
            \edge[ao(english),very thick] node[] {}; [\edge[ao(english),very thick]; [\edge[ao(english),very thick]; [\edge[ao(english),very thick]; [\edge[ao(english),very thick]; [.Spa ]]]]]
            ]
            \edge node[near end,right] {};
            [
            \edge node[near end,left] {};
            [
            \edge[ao(english),very thick] node[] {}; [\edge[ao(english),very thick]; [\edge[ao(english),very thick]; [\edge[ao(english),very thick]; [.Por ]]]]
            ]
            \edge node[near end,right] {};
            [
            \edge node[near end,left] {};
            [
            \edge[orange,very thick] node[] {}; [\edge[orange,very thick]; [\edge[orange,very thick]; [.Per ]]]
            ]
            \edge node[near end,right] {};
            [
            \edge node[near end,left] {};
            [
            \edge[ao(english),very thick] node[] {}; [\edge[ao(english),very thick]; [.Ita ]]
            ]
            \edge node[near end,right] {};
            [\edge node[near end,left] {};
            [
            \edge[ao(english),very thick] node[] {}; [.Fre ]
            ]
            \edge node[near end,right] {};
            [
            \edge[red,very thick] node[] {}; [.Ger ]
            ]]]]]]]
            \end{tikzpicture}
}
\end{minipage}
\begin{minipage}[b]{0.24\textwidth}
	\subfigure[Tree-to-string Patterns]{
\begin{tikzpicture}[every tree node/.style={font=\small},
            grow=left,
            level distance=0.45cm,sibling distance=.05cm,
            edge from parent path={(\tikzparentnode.south) -- +(-5pt,0) -| (\tikzchildnode)}],
            \Tree
            [
            \edge node[near end,left] {};
            [
            \edge[cyan,very thick] node[] {}; [\edge[cyan,very thick]; [\edge[cyan,very thick]; [\edge[cyan,very thick]; [\edge[cyan,very thick]; [\edge[cyan,very thick]; [\edge[cyan,very thick]; [.Hin ]]]]]]]
            ]
            \edge node[near end,right] {};
            [
            \edge node[near end,left] {};
            [
            \edge[red,very thick] node[] {}; [\edge[red,very thick]; [\edge[red,very thick]; [\edge[red,very thick]; [\edge[red,very thick]; [\edge[red,very thick]; [.Ger ]]]]]]
            ]
            \edge node[near end,right] {};
            [
            \edge node[near end,left] {};
            [
            \edge[orange,very thick] node[] {}; [\edge[orange,very thick]; [\edge[orange,very thick]; [\edge[orange,very thick]; [\edge[orange,very thick]; [.Per ]]]]]
            ]
            \edge node[near end,right] {};
            [
            \edge node[near end,left] {};
            [
            \edge[ao(english),very thick] node[near end,left] {};
            [
            \edge[ao(english),very thick] node[] {}; [\edge[ao(english),very thick]; [\edge[ao(english),very thick]; [.Ita ]]]
            ]
            \edge[ao(english),very thick] node[near end,right] {};
            [
            \edge[ao(english),very thick] node[near end,left] {};
            [
            \edge[ao(english),very thick] node[] {}; [\edge[ao(english),very thick]; [.Por ]]
            ]
            \edge[ao(english),very thick] node[near end,right] {};
            [
            \edge[ao(english),very thick] node[near end,left] {};
            [
            \edge[ao(english),very thick] node[] {}; [.Fre ]
            ]
            \edge[ao(english),very thick] node[near end,right] {};
            [
            \edge[ao(english),very thick] node[] {}; [.Spa ]
            ]] ] ]
            \edge node[near end,right] {};
            [
            \edge[blue,very thick] node[near end,left] {};
            [
            \edge[blue,very thick] node[] {}; [\edge[blue,very thick]; [\edge[blue,very thick]; [.Ukr ]]]
            ]
            \edge[blue,very thick] node[near end,right] {};
            [
            \edge[blue,very thick] node[near end,left] {};
            [
            \edge[blue,very thick] node[] {}; [\edge[blue,very thick]; [.Pol ]]
            ]
            \edge[blue,very thick] node[near end,right] {};
            [
            \edge[blue,very thick] node[] {}; [\edge[blue,very thick]; [.Rus ]]
            ] ] ]]]] ]
            \end{tikzpicture}
	}
\end{minipage}
\begin{minipage}[b]{0.24\textwidth}
\subfigure[Tree-to-tree Patterns]{
\begin{tikzpicture}[every tree node/.style={font=\small},
            grow=left,
            level distance=0.55cm,sibling distance=.05cm,
            edge from parent path={(\tikzparentnode.south) -- +(-5pt,0) -| (\tikzchildnode)}],
            \Tree
            [
            \edge node[near end,left] {};
            [
            \edge node[near end,left] {};
            [
            \edge node[near end,left] {};
            [
            \edge node[near end,left] {};
            [
            \edge[blue,very thick] node[] {}; [.Pol ]
            ]
            \edge node[near end,right] {};
            [
            \edge[blue,very thick] node[] {}; [.Ukr ]
            ]
            ]
            \edge node[near end,right] {};
            [
            \edge[blue,very thick] node[] {}; [\edge[blue,very thick];  [.Rus ] ]
            ]
            ]
            \edge node[near end,right] {};
            [
            \edge[red,very thick] node[] {};[\edge[red,very thick]; [\edge[red,very thick]; [.Ger ]]]
            ]
            ]
            \edge node[near end,right] {};
            [
            \edge node[near end,left] {};
            [
            \edge node[near end,left] {};
            [
            \edge node[near end,left] {};
            [
            \edge[ao(english),very thick] node[] {}; [.Spa ]
            ]
            \edge node[near end,right] {};
            [
            \edge[ao(english),very thick] node[] {}; [.Por ]
            ]
            ]
            \edge node[near end,right] {};
            [
            \edge node[near end,left] {};
            [
            \edge[ao(english),very thick] node[] {}; [.Ita ]
            ]
            \edge node[near end,right] {};
            [
            \edge[ao(english),very thick] node[] {}; [.Fre ]
            ]
            ]
            ]
            \edge node[near end,right] {};
            [
            \edge node[near end,left] {};
            [
            \edge[cyan,very thick] node[] {};[\edge[cyan,very thick]; [.Hin ]]
            ]
            \edge node[near end,right] {};
            [
            \edge[orange,very thick] node[] {};[\edge[orange,very thick]; [.Per ]]
            ]
            ]
            ]
            ]

            \end{tikzpicture}
	}
\end{minipage}
	\caption{The hierarchical clustering results on ten Indo-European languages (abbreviated as the first three letters) based on different features. Languages in the same genus are painted the same color.}
	\label{fig:typology_indo-enropean}
\end{figure*}
We reconstruct the family tree for 10 Indo-European languages\footnote{Germanic (German), Indic (Hindi), Iranian (Persian), Romance (Spanish, Italian, French, Brazilian Portuguese),Slavic (Russian, Polish, Ukrainian)}.
To compare with other features, we also implement clustering based on aligned word pairs and CFG rules extracted from non-native texts.
The produced dendrograms are shown in Figure \ref{fig:typology_indo-enropean} and numerical results are displayed in Table \ref{tab:res}.
We can see that simple CFG rules are not enough but delexicalized patterns can work as effectively as lexical information when predicting phylogenetic relationships.

%
%The first dendrogram is compatible with the widely recognised language family tree, which is not surprising, since lexical items is a main evidence of phylogenetic relatedness.
%Simply utilizing grammar rules extracted from automatically predicted L2 trees, as illustrated in the second dendrogram, can not produce satisfactory results.
%But we can reconstruct the phylogeny tree accurately with our proposed tree-string patterns, as illustrated in the last dendrogram.
%The resulting structure is slightly different from that based on lexical items whereas both dendrograms suggest that Hindi is farther to Romance and Slavic families than the other two single languages.
%Consistent results from different setups (See Appendices) strongly demonstrate the coherency between cross-lingual transfer and phylogeny.
We rank all the tree-string fragments by frequency in terms of each native language.
Statistical analysis of patterns renders interesting structures which are consistent with typical linguistic phenomena in certain languages, as shown in Figure \ref{fig:pattern}.
%Figure \ref{fig:pattern} shows several remarkably high-ranking patterns.
%The first pattern is the most frequent one in all 21 languages.
%The main contributing factor is the difficulty of distinguishing possessive and genitive forms (\textit{of}) in modern English.
%Explanations of the other three patterns correspond to the frequently-used compound structure in Mandarin, the incongruence of the position of adverbs between French and English, and the fixed word orders in interrogative sentences of Arabic.
Analyzing frequent cases of induced patterns, we find that they reveal the information about how the delexicalized knowledge of native languages is transferred into the second language, which can benefit the understanding of interlanguage and second language acquisition (SLA).
\section{Conclusion}
In this paper, we investigate to what extent does a single delexicalized feature reflect the influence from traditional phylogenetic language groups.
In particular, we design two kinds of interpretable patterns and induce them from parallel data with grammar induction technologies.
The produced family tree of 10 Indo-European languages match well with the phylogenetic structure resulting from historical-comparative paradigm.
It suggests the coherence between genetic relationships and delexicalized cross-linguistic transfer, and more experiments need to be implemented in future research.

%
%In this paper, we claim that the divergence patterns formulated by historical linguists and typologists reflect constraints on human languages, and are thus consistent with SLA.
%We validate this hypothesis on large-scale text data from English learners of ten representative Indo-European languages.
%We formalize the transfer of knowledge from L1 to L2 as interpretable tree-string patterns in the framework of STSG.
%% Such patterns can be automatically induced from web data by applying state-of-the-art syntactic parsing and grammar induction technologies.
%% This allows us to quantitatively probe cross-lingual transfer and extend inquiries of SLA.
%We present, for the first time, an empirical evidence to support the agreement between cross-lingual transfer and the phylogenetic structure resulting from the historical-comparative paradigm.

\bibliographystyle{coling}
\bibliography{coling2020}

\end{document}